\def\year{2022}\relax
\documentclass[letterpaper]{article} 
\usepackage{aaai22}  
\usepackage{times}  
\usepackage{helvet}  
\usepackage{courier}  
\usepackage[hyphens]{url}  
\usepackage{graphicx} 
\usepackage{natbib}  
\usepackage{caption} 
\DeclareCaptionStyle{ruled}{labelfont=normalfont,labelsep=colon,strut=off} 
\usepackage{algorithm}
\usepackage{algorithmic}
\usepackage{multirow}
\usepackage{booktabs}
\usepackage{newfloat}
\usepackage{listings}
\floatstyle{ruled}
\newfloat{listing}{tb}{lst}{}
\floatname{listing}{Listing}
\usepackage{amsmath}
\usepackage{amsfonts}
\usepackage{mathtools}
\title{Energy Attack: On Transferring Adversarial Examples}
\author{
    Ruoxi Shi\equalcontrib, Borui Yang\equalcontrib, Yangzhou Jiang, Chenglong Zhao, Bingbing Ni\thanks{Corresponding author.}
}
\affiliations{
    Shanghai Jiao Tong University \\
    \{eliphat, ybirua, jiangyangzhou, cl-zhao, nibingbing\}@sjtu.edu.cn

}

\usepackage{bibentry}

\begin{document}

\maketitle

\begin{abstract}
In this work we propose Energy Attack, a transfer-based black-box $L_\infty$-adversarial attack. The attack is parameter-free and does not require gradient approximation. In particular, we first obtain white-box adversarial perturbations of a surrogate model and divide these perturbations into small patches. Then we extract the unit component vectors and eigenvalues of these patches with principal component analysis (PCA). Base on the eigenvalues, we can model the energy distribution of adversarial perturbations. We then perform black-box attacks by sampling from the perturbation patches according to their energy distribution, and tiling the sampled patches to form a full-size adversarial perturbation. This can be done without the available access to victim models. Extensive experiments well demonstrate that the proposed Energy Attack achieves state-of-the-art performance in black-box attacks on various models and several datasets. Moreover, the extracted distribution is able to transfer among different model architectures and different datasets, and is therefore intrinsic to vision architectures.

\end{abstract}

\section{Introduction}
With the wide application of artificial neural networks comes a growing research interest into the robustness of neural network models \cite{carlini2019evaluating,xu2020adversarial}, in the manner of adversarial attacks: neural networks can be fooled to produce incorrect outputs by slightly perturbed inputs known as adversarial examples \cite{szegedy2014intriguing,biggio2013evasion}. There are in general two classes of threat models\footnote{A threat model defines what the attacker knows about the victim.}, and the respective classes of attacks: white-box threat models/attacks and black-box threat models/attacks, which assume different knowledge about the target neural networks.

White-box attacks assume full access to the target model and usually require the first-order (i.e., gradient) information of the target model \cite{szegedy2014intriguing, goodfellow2015explaining, kurakin2016adversarial, moosavi2016deepfool}. While white-box attacks are generally more powerful, they rely on direct access to the target model to compute the gradients.
On the other hand, black-box attacks only assume access to the inputs and outputs of the target model. Therefore, in many real{-}world scenarios, where full access to {the} target models may not {be} available, black-box attacks are considered more practically feasible. For black-box attacks, gradient approximation and random search \cite{rastrigin1963convergence} are two frequently applied schemes. The former approximates the gradient of {a} black-box model by querying the model directly \cite{chen2017zoo, tu2019autozoom}, or by transferring gradient information from a white-box surrogate model \cite{cheng2019improving}, and the latter simply performs greedy search steps at random. However, there are two major challenges: first, it may take an enormous number of queries for a black-box method to successfully find an adversarial example, as {the} gradient approximation is non-trivial in high dimension \cite{ilyas2019prior}; second, the attack success rate of these methods are usually lower than white-box attacks \cite{andriushchenko2020square}. A recent work \cite{andriushchenko2020square} copes with these challenges with Square Attack, a random-search-based method that samples perturbations from a specific distribution.

In this work, we propose Energy Attack, a query-efficient transfer-based black-box attack that also applies the random search framework. Notice that in {our} random search scheme, finding a proper distribution to sample from is one of the keys to improving the success rate and query efficiency. Different from the classic random search algorithm, which samples perturbations uniformly at random, and Square Attack, which samples homogeneously colored squares as perturbations, Energy Attack samples perturbations based on the information obtained from white-box perturbations of a surrogate model.

The sampling strategy of Energy Attack is based on our observation that the patches (i.e., small square regions) of white-box perturbations share similar energy distributions on several unit directions, even if the perturbations themselves come from different models trained on different datasets. Especially, perturbation patches of robust models can better capture the shared distribution among patches of different models.
We may extract such distributions from a surrogate model and draw perturbations from them when attacking. This allows exploiting the similarities among perturbation patches, boosting the black-box attack based on random search.

Therefore, Energy Attack first extracts a set of perturbation patches and corresponding energy distributions by performing principal component analysis (PCA) on white-box perturbation patches of a surrogate model. Then in the actual black-box attack phase, it samples from the extracted patches according to the energy distribution, and tiles the patches to form a larger square region of perturbation. If adding this perturbation increases the loss of the target model, Energy Attack will accept the change, otherwise {the change will be discarded}.

Unlike previous transfer-based attacks, which mainly aims at estimating gradients from surrogate models \cite{cheng2019improving}, Energy Attack directly transfers extracted perturbations, in the form of their energy distributions in patches. We highlight that the white-box surrogate model is only used to acquire white-box perturbations and we do not need the presence of the white-box model in the black-box attack phase, which makes the attack phase more efficient. Also, we will demonstrate in Section\ \ref{sec:Experiments} that such distributions generalize across model architectures and image domains, thus we do not require the surrogate model to have {the} same training source or similar structures as the target model.

\begin{figure*}[t]
\centering
\includegraphics[width=0.75\textwidth]{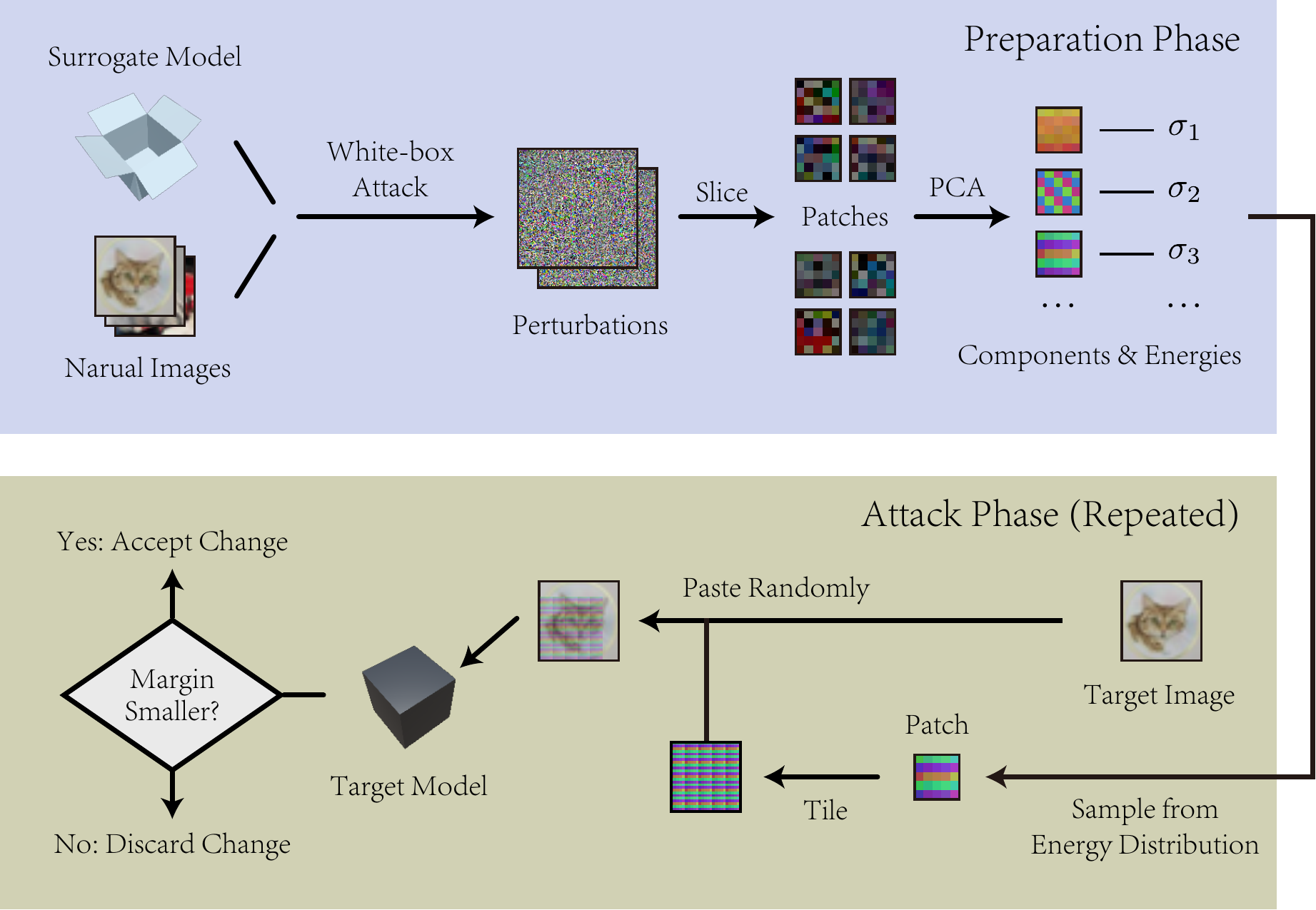}
\caption{\textbf{Pipeline of Energy Attack.} Energy Attack is a transfer-based attack, and has two phases. In the preparation phase, we collect patches from the perturbations of a white-box attack on a surrogate model, and perform PCA on them to obtain the energy distribution in patches. In the attack phase, we sample patches from the energy distribution, tile them up and perform a random search procedure.}
\label{fig:fancy}
\end{figure*}

\section{Related Work}
\paragraph{Robustness of White-Box Neural Networks} It was first discovered by Szegedy et al. \cite{szegedy2014intriguing} that naturally trained neural networks are not robust under certain small perturbations. Since then, adversarial attacks and defenses of neural networks under a fully informed white-box scenario have been widely proposed \cite{athalye2018obfuscated,moosavi2016deepfool,pang2020boosting,zhang2019theoretically}.

Recently, progress was made on parameter-free adversarial attacks by the framework proposed by Chen et al. \cite{chen2020frank}, based on the reduced gradient Frank-Wolfe algorithm \cite{frank1956algorithm} for constrained convex optimization. Furthermore, white-box attacks have not only been strong evaluators for robustness, but also a tool for analyzing other behaviours of neural networks. For example, robust networks own visually aligned gradients \cite{kaur2019perceptuallyaligned} and invertible feature spaces \cite{engstrom2019adversarial}, which are also desired properties in addition to robustness itself \cite{salman2020adversarially}.

\paragraph{Black-Box Attacks} While the white-box scenario is not quite common in real life, black-box attacks only require classification results or score information and can practically show the underlying non-robustness of real-world vision models \cite{ilyas2018black,chen2017zoo}. Early approaches under this scenario mainly focus on approximating the gradients of target models \cite{chen2017zoo,ilyas2018black,guo2019simple}, but they suffer from large number of queries and low attack success rates. For better performance, Bandits \cite{ilyas2019prior}, SignHunter \cite{aldujaili2019bit} and Square Attack \cite{andriushchenko2020square} employed different strategies to achieve higher success rates as well as higher query efficiencies. Square Attack \cite{andriushchenko2020square} also applies a random search \cite{rastrigin1963convergence} procedure instead of stochastic approximation. Others employ transfer-based priors.

\paragraph{Transferability of Adversarial Examples and Transfer-Based Attacks} Szegedy et al. \cite{szegedy2014intriguing} first finds that some adversarial examples can generalize to different neural networks. Moosavi et al. \cite{moosavi2017universal} further extended it to the universal case where the same perturbation is used to fool all models. An intriguing fact is that some of the examples even `transfer' to time-limited humans \cite{elsayed2018adversarial}, showing that common intrinsic properties exist in vision tasks. Based on this transferability, people have achieved success in highly efficient attacks \cite{yan2019subspace,cheng2019improving,dolatabadi2020advflow,huang2019black} in the black-box scenario.

\section{Energy Attack}\label{sec:EA}
Now we introduce the proposed Energy Attack in detail. The algorithm has two phases: a preparation phase and an attack phase. The pipeline of Energy Attack is visualized in Figure\ \ref{fig:fancy}.

\subsubsection{Preparation Phase} In the preparation phase, we first attack a surrogate model with a white-box {method}. In particular, we apply the Frank-Wolfe framework proposed in \cite{chen2020frank} for parameter-free white-box attacks, though any effective attack should work basically the same. We collect all small patches of a specific size ($s_p \times s_p$) in the examples generated by the attack, and perform PCA on {them}. Notice that no matter how many patches there are, the co-variance matrix remains to be low-dimensional with $cs_p^2$ rows and columns where $c$ is the number of channels. Moreover, the matrix can be computed on-the-fly when performing the white-box attack, without storing full results in the memory. Thus the procedure is quite efficient.

PCA yields a set of energy values $\sigma_i$ and the corresponding orthonormal eigenbasis $u_i$. Since the PCA is performed on a set of $c \times s_p \times s_p$ patches, each vector in the eigenbasis actually correspond to a small patch in the image space.

Formally, we fetch the tensor formed by {the} patches of adversarial examples as
\begin{equation}
    \mathbf{P} \in \mathbb{R}^{N \times c \times s_p \times s_p},
\end{equation}
and flatten the latter three dimensions of $\mathbf{P}$ into matrix
\begin{equation}
    \mathbf{\tilde{P}} \in \mathbb{R}^{N \times cs_p^2}.
\end{equation}
The co-variance matrix is then computed by
\begin{equation}
    \mathbf{K} =\mathbf{\tilde{P}}^T\mathbf{\tilde{P}} \in \mathbb{R}^{cs_p^2 \times cs_p^2},
\end{equation}
on which PCA is subsequently performed, as an eigenvalue decomposition of Hermitian $\mathbf{K}$:
\begin{equation}
    \mathbf{K} = \mathbf{U}\Sigma\mathbf{U}^{-1},
\end{equation}
where $\mathbf{U} = (u_1, u_2, \dots, u_n)$ is an orthonormal basis and $\Sigma$ is a diagonal matrix formed from corresponding energy values $\langle \sigma_i \rangle$.

We utilize these results in the attack phase to boost the black-box attack.

Different from previous transfer-based approaches such as \cite{huang2019black} and \cite{dolatabadi2020advflow}, since we operate on the level of patch, the surrogate model {is not required} to accept images of the same size as the target model, nor be trained on the same dataset. Actually, we demonstrate that the distribution of patches captured by PCA transfers well between different datasets and different model architectures later in Section\ \ref{sec:Experiments}.

\subsubsection{Attack Phase} In {the} attack phase, we apply a random search \cite{rastrigin1963convergence} procedure to find adversarial examples. The algorithm for the attack is illustrated in Alg.\ \ref{alg:EnergyAttack}.

\begin{algorithm}[!b]
\caption{The Energy Attack (Attack Phase)}
\label{alg:EnergyAttack}
\textbf{Input:} Target loss $\ell$ to maximize, image $x$, $L_\infty$ bound $\epsilon$, PCA results $\{\sigma_i\}$ and $\{u_i\}$. \\
\textbf{Output:} Perturbation $\delta$. \\
\textbf{Notation:} $s_x$ is the side length of image and $s_p$ is the side length of PCA patches.

\begin{algorithmic}[1]
\STATE $\delta \leftarrow \verb+random_stripes()+, \ell^* \leftarrow -\infty$
\STATE $t \leftarrow \lfloor s_x / s_p \rfloor$
\FOR{maximum $n$ queries}
\STATE Sample $v = u_i$ with probability $\sigma_i/\sum_j{\sigma_j}$
\IF{$t > 0$}
\STATE $V \leftarrow$ Tile $v$ by $t$ times on each side
\ELSE
\STATE $V \leftarrow$ Randomly pick $3 \times 3$ patch from $v$
\ENDIF
\STATE $P \leftarrow \epsilon \cdot \verb+random_sign()+ \cdot \verb+sign(+ V \verb+)+$
\STATE $\delta' \leftarrow$ Randomly replace a $ts_p \times ts_p$ area in $\delta$ with $P$
\STATE \algorithmicif\ {$\ell(\delta') > \ell^*$} \algorithmicthen\ $ \delta \leftarrow \delta', \ell^* \leftarrow \ell(\delta')$
\STATE \algorithmicif\ $\verb+hopeless()+$ \algorithmicthen\ $t \leftarrow \lfloor t / 2 \rfloor$
\ENDFOR
\end{algorithmic}
\end{algorithm}

For initialization, we follow \cite{andriushchenko2020square} to initialize the perturbation with randomly colored vertical stripes.

For each step, we first sample a patch from the PCA eigenbasis $u_i$, each vector (or equivalently, image patches) with probability proportional to $\sigma_i$. Formally, the distribution is

\begin{align}
\begin{split}
    &\mu = (\Omega, 2^\Omega, P), \\
\end{split}
\end{align}

where
\begin{align}
\begin{split}
    &\Omega = \{u_1, u_2, \dots, u_n\} \\
    &P(u_i) = \frac{\sigma_i}{\sum_{j=1}^n \sigma_j}, \forall i \in \{1, 2, \dots, n\} \\
\end{split}
.
\end{align}

\begin{table*}[t]
\centering
\begin{tabular}{@{}ccccccccccccc@{}}
\toprule
\multirow{2}{*}{}     & \multicolumn{3}{c}{VGG16BN}              & \multicolumn{3}{c}{InceptionV3}           & \multicolumn{3}{c}{ResNet18}              & \multicolumn{3}{c}{ViTB16}                  \\
                      & Avg.       & Med.       & ASR            & Avg.        & Med.       & ASR            & Avg.        & Med.       & ASR            & Avg.         & Med.        & ASR            \\ \midrule
$\text{Bandits}_{\mathit{TD}}$ & 370        & 90         & 91.6           & 1117        & 140        & 95.4           & 168         & 92         & 91.7           & -            & -           & -              \\
P-RGF   & 557        & 82         & 95.9           & 694         & 80         & 86.9           & 411         & 62         & 97.9           & -            & -           & -              \\
Square                & 24         & \textbf{1} & \textbf{100.0} & 207         & 45         & 99.5           & 34          & 6          & \textbf{100.0} & \textbf{125} & 36          & \textbf{100.0} \\
Subspace              & 104        & 18         & \textbf{100.0} & 174         & 21         & 99.7           & 51          & 14         & \textbf{100.0} & 344          & 108         & 99.7           \\
AdvFlow               & 929        & 400        & 99.2           & 1081        & 400        & 97.5           & 528         & 400        & 99.5           & 1057         & 400         & 98.4           \\
TREMBA                & \textbf{6} & \textbf{1} & \textbf{100.0} & 264         & 22         & 99.9           & 27          & \textbf{1} & \textbf{100.0} & 174          & \textbf{22} & 99.9           \\ \midrule
Energy (R)            & 13         & \textbf{1} & \textbf{100.0} & \textbf{91} & \textbf{1} & \textbf{100.0} & \textbf{20} & \textbf{1} & \textbf{100.0} & \textbf{125} & 42          & \textbf{100.0} \\
Energy (NR)           & 18         & \textbf{1} & \textbf{100.0} & 201         & 4          & 99.9           & 26          & \textbf{1} & \textbf{100.0} & 225          & 61          & 99.8           \\ \bottomrule
\end{tabular}
\caption{\textbf{Attack success rate and query efficiency of attacks against different ImageNet models.} (R) denotes Energy Attack using PCA results of a robust model, and (NR) denotes the attack using PCA results of a non-robust model. Our method achieves outstanding results on this standard black-box attacking benchmark.}  
\label{tab:ImageNetAttackResults}
\end{table*}

We sample from this distribution because if we sample many patches according to it and perform the PCA again, we will still result in the same distribution. The patch is then either tiled into a larger rectangular area of perturbations, or further cropped into a smaller one according to the stage of the attack. We enforce the $L_\infty$ constraints, and replace a random area with {an} appropriate size in the current best-scored perturbation with the tiled or cropped area.

Then we take the random search procedure: we keep the new perturbation in record if the resulting new perturbation is better in score, and discard otherwise. In case the algorithm figures out that there is little hope advancing with the current tiling/cropping configuration, we half the number of tiling $t$. For briefness we denote the situation as `hopeless' in the following discussion. The decision criteria {is} also referred to as tiling strategies later, since this essentially determines how Energy Attack tiles the patches.

If there is a batch of images to attack, we consider the state hopeless if no loss increment is seen in current step for all images. If we have only one image to attack, we use a probabilistic model -- we consider it hopeless if with $90\%$ confidence the probability of advancing for each step is less than $15\%$, which can be inferred from 15 consecutive steps without loss increment. In our experiments we only show results for the first procedure since we find that they behave similarly (see Section\ \ref{sub:ValidatingEA}) and thus it is merely a matter of code efficiency.

For the target loss, we employ the standard margin loss proposed in \cite{andriushchenko2020square}:
\begin{equation}
    \ell(y, \hat{y}) = \hat{y}_{\pi2} - \hat{y}_{\pi1},
\end{equation}
where $\pi$ denotes the permutation that sorts the logits in descending order. The margin loss is the difference between the logit with the largest and the second largest value. We also filter the target images that we already succeed in attacking so that the margin loss faithfully reflects the margin to a wrong classification, as in previous work \cite{andriushchenko2020square}.

\section{Results}\label{sec:Experiments}
In this section, we conduct extensive experiments to evaluate the effectiveness and transferability of Energy Attack.

\subsection{Benchmarking Energy Attack}\label{sub:BenchmarkingEA}

\subsubsection{Experiment Setup}
We benchmark Energy Attack on several models, using 1000 images randomly chosen from ImageNet \cite{deng2009imagenet} validation sets. Apart from Energy Attack, six other black-box attacks are also included for parallel comparison: 1) Bandits \cite{ilyas2019prior}, 2) P-RGF with $\lambda^*$ configuration \cite{cheng2019improving}, 3) Square Attack \cite{andriushchenko2020square} 4) Subspace Attack \cite{yan2019subspace}, 5) AdvFlow \cite{dolatabadi2020advflow} and 6) TREMBA \cite{huang2019black}. Models used for benchmarking include batch normalized VGG16 (VGG16BN) \cite{simonyan2014very}, Inception V3 \cite{szegedy2016rethinking}, ResNet18 \cite{he2016deep} and ViTB16 \cite{dosovitskiy2020image}. For Energy Attack, we use PCA results from a non-robust ResNet34 model and an adversarial-trained robust ResNet18 model respectively. Both surrogate models are originally trained on the CIFAR10 \cite{krizhevsky2009learning} dataset. Other attacks are configured to their recommended parameters on ImageNet. All attacks are restricted by a maximum $L_{\infty}$ distance of $\epsilon=0.05$, and a maximum of 10000 queries. The models and Energy Attack are implemented in PyTorch \cite{paszke2017automatic}.

\begin{figure*}[t]
\centering
\includegraphics[width=0.85\textwidth]{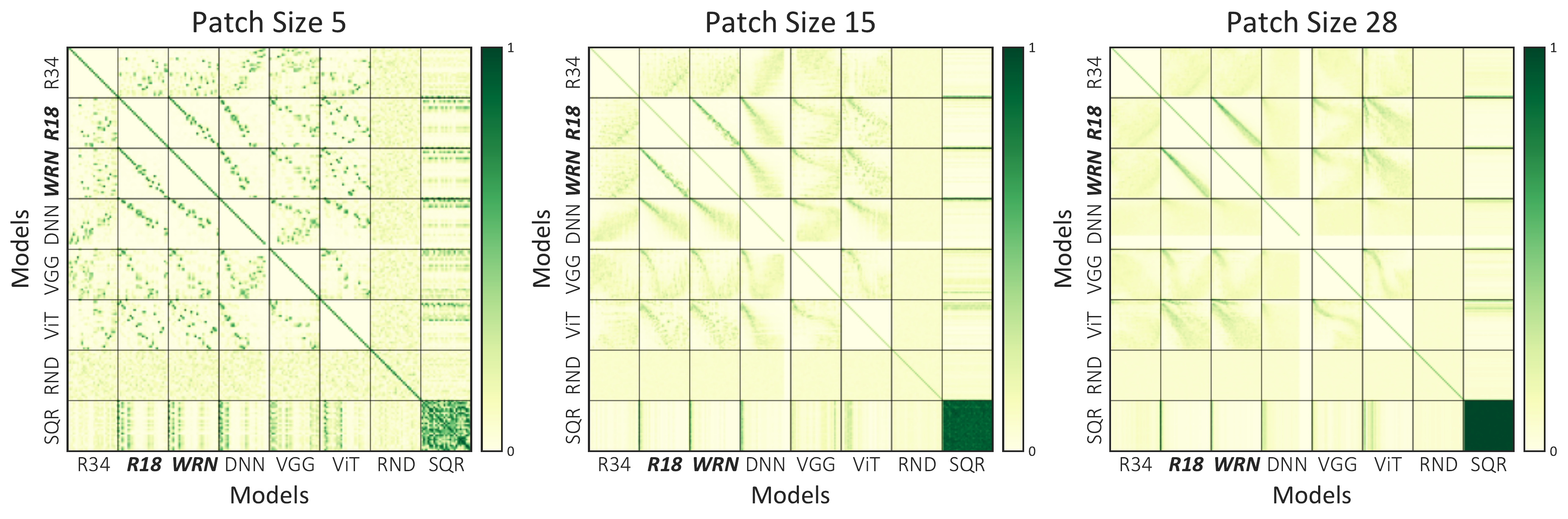}
\caption{\textbf{Pairwise cosine similarity of patches from models on different datasets.} We use ResNet34 (R34), ResNet18 (R18) and WideResNet28 (WRN) from CIFAR, DNN (DNN) from MNIST, and VGG16BN (VGG) and ViTB16 (ViT) from ImageNet. We also include patches of uniformly random images (RND) and random colored squares (SQR). Model names in bold and italic represent robust models. We choose the first one-third of total patches from the PCA results, and sort the patches by their corresponding eigenvalues in descending order. If the PCA result contains fewer effective vectors (due to mathematically or numerically low-rankness) than necessary, we pad with zero vectors.}
\label{fig:TriDotProduct}
\end{figure*}

\subsubsection{Attack Results}
Table \ref{tab:ImageNetAttackResults} reports the attack results against the four models. In most cases, Energy Attack achieves performances equal to or even better than existing state-of-the-art black-box attacks, in terms of average and median number of queries and attack success rate. Energy Attack with robust perturbation patches achieves 100.00\% attack success rate against all four models in the experiment, and three out of these four attacks take the least average queries. This achievement outperforms all four other black-box attacks used in this experiment, which demonstrates the outstanding performance of our proposed method.

Notice that for Energy Attack, the target models, with various architectures, are trained on ImageNet, but we are using the PCA patches derived from CIFAR10 ResNet models to perform the attack. TREMBA uses a hierarchical convolutional encoder stack which is similar to the structure of VGG networks. Subspace attack uses three ResNets as the surrogate models. Thus they achieved higher-than-average performances on these specific architectures, but are weaker on others. Our method captures more intrinsic nature of images, and generalizes well cross {various} architectures and image domains.


\subsection{When Do the Patches Transfer?}\label{sub:TransferAttackInsights}

In this section, we conduct analysis on the similarities among patches of different sizes extracted from different models and different datasets, in order to gain a deeper insight into the transferability of perturbation patches.

\subsubsection{Sizes: Localized Distributions Transfer Better}
We first consider the choice of patch sizes. We select 6 models from the CIFAR10, MNIST \cite{lecun1998gradient} and ImageNet datasets: ResNet34, ResNet18, WideResNet28 from CIFAR10, DNN from MNIST, and VGG16-BN and ViTB16 from ImageNet. Among them the ResNet18 and WideResNet28 are robust models, with the former trained with standard PGD-10 Adversarial Training \cite{madry2017towards} and the latter trained with TRADES \cite{zhang2019theoretically}. We extract PCA patches of different sizes from white-box adversarial perturbations for each model and sort them by corresponding energy values in descending order. In particular, we choose patch sizes to be 5, 15 and 28, from a small size to the scale of full image for MNIST and CIFAR. We then select the first one-third patches with highest energy concentration for each model and compute the cosine similarity for each pair of patches. Note that patches of the first two models (ResNet34 and ResNet18) are exactly the two sets of patches used in Section \ref{sub:BenchmarkingEA}.

Figure \ref{fig:TriDotProduct} illustrates the result of all pairwise cosine similarities. The cosine similarity between different models vanishes (degrades to the level of similarity with uniformly random samples) as {the} patch size increases. This explains why previous approaches that try to capture the perturbation distribution of an entire image do not transfer well. The patch sizes in these approaches are too large to retain the similarities among different models, which makes the transferred attacks less effective. On the other hand, smaller patches that focus on the localized structure of perturbations transfer better, as they capture local perturbation distributions that share more similarities with perturbation distributions of other models. 

It is also worth noting that the pure-colored squares also have strong similarities with patches of some models in our experiment, which supports the design of square-region perturbations in Square Attack \cite{andriushchenko2020square}. Their design originally comes from the analysis on convolutional operators.


\subsubsection{Sources: Robust vs. Non-Robust Models}
Patches from robust models transfer better to other models. In Table \ref{tab:ImageNetAttackResults}, Energy Attack using robust patches outperforms the attack using non-robust patches in all three measurements. Energy Attack with robust patches takes fewer queries and has a higher success rate against all four models in the benchmark. Using robust patches, it achieves 100\% success rate against all four models, while the same attack using non-robust patches is only able to achieve 100\% success rate against two of the models. Additionally, the numbers of average queries and median queries of Energy Attack with robust patches are also smaller than those of Energy Attack with non-robust patches.

These experiment results are in accordance with the visualization of pairwise cosine similarities illustrated in Figure \ref{fig:TriDotProduct}. Compared with non robust models, robust models have larger cosine similarities with PCA patches from other models. When we transfer non-robust patches to other models (i.e., compute the pairwise cosine similarities between patches of non-robust models and other models), the values of cosine similarity tend to scatter across the entire heatmap. However, when we transfer robust patches to other models, the values of cosine similarities tend to be distributed along the diagonal. This indicates that, compared to patches from non-robust models, patches from robust models can better capture the underlying perturbation distributions shared among most models.

Another interesting observation is that patches from the two robust models (ResNet18-AT and WideResNet28-TRADES) have generally the same distribution. The values of cosine similarities between these two models concentrate on the diagonal of the heatmap, indicating that the patches are virtually orthogonal, even though they come from two different models.

\subsubsection{Cosine Similarity as a Measurement}
Based on discussions in previous subsections, we propose that the cosine similarity can be used as a measurement to indicate the transferability of a set of extracted patches. This conclusion follows from the observation that the performances of Energy Attack reported in Table \ref{tab:ImageNetAttackResults} correspond to the visualization results in Figure \ref{fig:TriDotProduct}. Recall that we use the patches of a non-robust ResNet34 and a robust ResNet18 on CIFAR10 (whose patches corresponds to R18 and R34 in Figure \ref{fig:TriDotProduct}, respectively) to attack models on ImageNet in Table \ref{tab:ImageNetAttackResults}. Both attacks achieve outstanding performance against VGG16BN, with 100\% ASR and median query of 1. Corresponding to this result, the cosine similarities in Figure \ref{fig:TriDotProduct} reflect similarities between distributions of energy in patches of CIFAR10 ResNet18, ResNet34 and ImageNet VGG16BN. Similarly, both attacks have a slight drop in performance when attacking ViTB16. While non-robust patches achieve 99.8\% ASR with an average of 225 queries, robust patches achieve better performance, with 100\% ASR and 125 queries on average. This result is also consistent with the visualization: when transferred to ViTB16, the patches of robust ResNet18 have smaller distribution shift, as the cosine similarities still tend to lie around the diagonal. In contrast, the patches of non-robust ResNet34 have larger distribution shift, corresponding to its poorer performance against ViTB16. Therefore, we can use the cosine similarity as a qualitative indicator for the transferability of PCA patches.

\begin{figure}[t]
\centering
\includegraphics[width=0.65\linewidth]{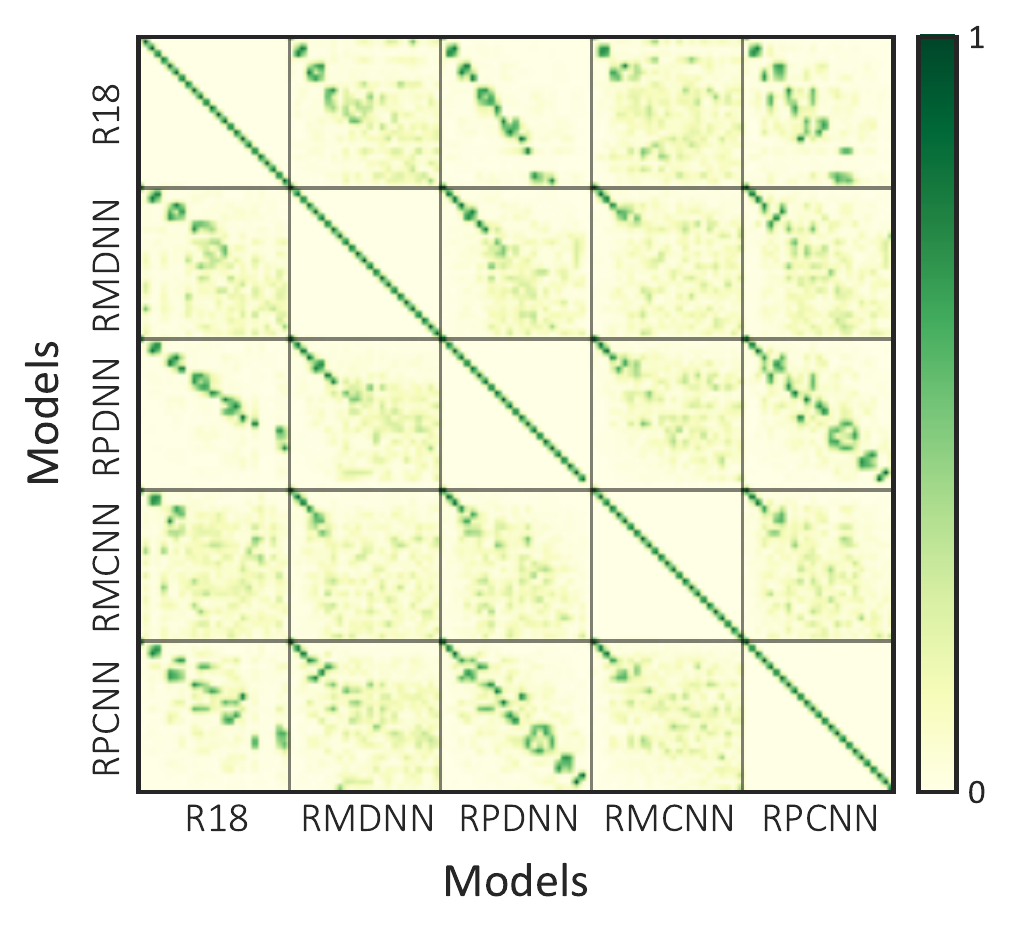}
\caption{\textbf{Pairwise cosine similarity of patches from various configurations.} The pairwise cosine similarity include a pair of randomly initialized, untrained MLP (RMDNN) and LeNet (RMCNN) and another pair of MLP and LeNet trained on MNIST with randomly permuted labels (RPDNN and RPCNN respectively). As a reference, patches from CIFAR10 robust ResNet18 are also included.}
\label{fig:SmallerDotProd}
\end{figure}



\subsubsection{Labels Are Not Important}
To help find out what caused this transferring behaviour, we further perform the following experiment. We run the preparation phase of Energy Attack (i.e., extract white-box perturbation patches and perform PCA) on two multi-layer perceptron (MLP) models and two LeNet models on the MNIST dataset. One pair of MLP and LeNet are randomly initialized and untrained, and another pair of MLP and LeNet are trained on MNIST with randomly assigned labels. For these models, the attack is run to make the model diverge from its decisions on clean images, instead of ground truths. We then compute the pairwise cosine similarity using the patches of these four models. The patches from the CIFAR10 robust ResNet18 are also included for reference.

The heatmap of cosine similarities is illustrated in Figure \ref{fig:SmallerDotProd}. It can be observed that the patches from random-initialized MLP and LeNet models do not share much similarity. However, in the case of trained models, even though the labels are randomly assigned, the training process still induces similarities in perturbation patches, as higher values now tend to concentrate along the diagonal.

\begin{table}[!b]
\centering
\begin{tabular}{@{}ccccc@{}}
\toprule
             & \multicolumn{2}{c}{ResNet34}  & \multicolumn{2}{c}{LeNet}     \\
             & Avg.        & ASR             & Avg.        & ASR             \\ \midrule
Square       & 566         & 99.95           & 123         & \textbf{100.00} \\
Energy (R)   & 88          & 99.99           & 36          & \textbf{100.00} \\
Energy (Res) & \textbf{50} & \textbf{100.00} & 129         & \textbf{100.00} \\
Energy (Le)  & 98          & 99.95           & \textbf{27} & \textbf{100.00} \\ \bottomrule
\end{tabular}
\caption{\textbf{Attack success rate and average queries of model patches against model themselves.} We use Energy Attack to attack a ResNet34 on CIFAR10 and a LeNet on MNIST, using PCA patches from the ResNet34 (Res) and the LeNet (Le). We also include attack results of Square Attack and Energy Attack (R) for reference.}
\label{tab:SelfAttack}
\end{table}

More interestingly, the patch distributions of randomly-trained MNIST MLP and LeNet also share similarities with the adversarially trained CIFAR10 ResNet18. Since these patches are derived from the small perturbations which cause the models to make wrong predictions, it follows that the perturbation directions of these patches are generally those achieving the strongest activations on artificial neural units. Thus, the similarities in PCA results indicate that the patches achieving high activation values on their respective models and datasets are essentially similar, regardless of datasets and labels. We thus conclude that these patches are independent of datasets and are intrinsic to vision architectures.

This conclusion is in accordance with the work of \cite{maennel2020neural}, which discovers that pre-training neural networks on image datasets with randomly assigned labels sometimes has positive impact on downstream tasks: the models might have captured these intrinsics during the pre-training process. It also agrees with the better transferability of robust neural models discovered by \cite{salman2020adversarially}: if we consider the neural units of different models activated by similar patches to have similar `functionalities' (i.e., they capture a similar set of features in input), then robust models, whose neural units are able to better capture the shared features among all image datasets, naturally transfer better for cross-domain image tasks.


\subsection{Verification of Designs}\label{sub:ValidatingEA}

\subsubsection{Power of Energy Distribution of Patches}

We first show that the energy distribution of patches is indeed an effective prior for transferred black-box attacks. We run the preparation phase of Energy Attack on CIFAR10 ResNet34 and a MNIST LeNet5 to extract their energy distribution of patches respectively. Then we use the patches to perform Energy Attack against them. The results of Square Attack and Energy Attack with robust patches are also included for comparison.

Table \ref{tab:SelfAttack} reports the success rate and {the number of} average queries. Energy Attack (R) reaches good performance on both models; when we attack the models using the distribution from the models themselves, the performance of Energy Attack is even better. However, since energy distribution of patches from non-robust models does not generalize as well as that from robust ones, patches from ResNet34 and LeNet achieves worse performance than the robust baseline when attacking each other.

The performance of Energy Attack in this `self-attack' setting shows that the PCA procedure indeed capture (at least partially) the distribution of the original white-box perturbations: using the energy distribution of a certain model against itself boosts performance of Energy Attack. Thus the energy distribution serves as an effective prior for attacking models in black-box scenarios.
\begin{table}[!b]
\centering
\begin{tabular}{@{}cccc@{}}
\toprule
            & VGG16BN & ResNet18 & InceptionV3 \\ \midrule
Probability & 13      & 17       & 94          \\
Batch       & 13      & 20       & 91         \\ \bottomrule
\end{tabular}
\caption{\textbf{Mean queries of two tiling strategies on ImageNet models.} The batch-based strategy and the probability-based strategy have similar performance when attacking the models on ImageNet.}
\label{tab:TilingStrategy}
\end{table}

\subsubsection{Tiling Strategies}
In Section \ref{sec:EA}, we proposed two schemes for adjusting tiling sizes during the attack: a batch-based method and a probability-based method. For the batch-based one, we reduce tiling sizes when none of perturbed images in the current batch increase {the} model prediction loss. The latter is for {the purpose of} decoupling all relations between target images, or attacking one image at a time. We reduce tiling sizes when no increment in loss is seen for 15 consecutive steps. We now show that the two strategies have similar performances. We use Energy Attack with robust model patches to attack three classic ImageNet models: VGG16BN, ResNet18 and InceptionV3. We perform two rounds of attack, using the probability-based scheme and the batch-based scheme respectively, and report the average number of queries.

Table \ref{tab:TilingStrategy} reports the average queries of the two strategies. The performances of these two strategies are very close, thus we pick the batch-based one because of its efficiency.

\subsubsection{Decision Threshold of `Hopeless'}
Energy Attack considers a tiling strategy `hopeless' if no image in the current batch has increment in loss {for $\tau$} consecutive perturbation steps. In our implementation, we set the threshold $\tau$ to one step. Now we configure Energy Attack to use different numbers of perturbation steps to decide whether the current tiling is `hopeless'. 
In particular, we vary the step threshold $\tau$ and run Energy Attack on the MNIST LeNet and the ImageNet Inception V3. We report the average queries of Energy Attack in each configuration.

\begin{figure}[t]
\centering
\includegraphics[width=0.99\linewidth]{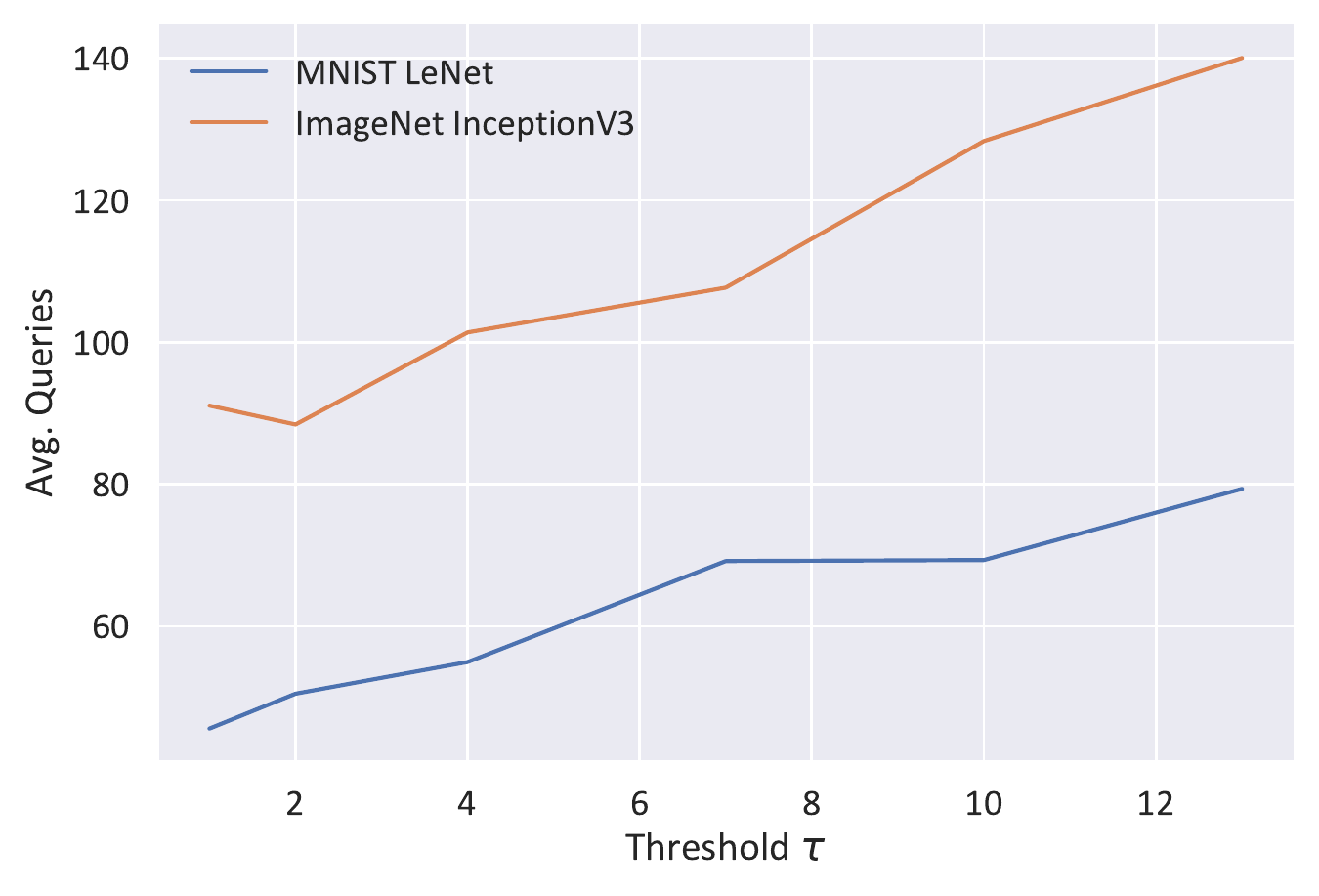}
\caption{\textbf{Average queries with regard to threshold $\tau$.} We set the threshold to different values, run Energy Attack on an MNIST LeNet and an ImageNet Inception V3 and plot the average queries under different threshold.}
\label{fig:BatchTresholdLinePlot}
\end{figure}

Figure \ref{fig:BatchTresholdLinePlot} plots the change in {the number of} average queries with regard to the threshold $\tau$. As the threshold increases, both models spend more queries on average. Therefore, using thresholds {larger} than one may instead exert negative effects on performance. Also, the trends of {the number of} average queries growing with threshold $\tau$ are similar, although they have different architectures and are on different datasets. Therefore this hyper-parameter does not need to be specifically tuned when attacking different models, and the entire procedure of Energy Attack is parameter-free.

\section{Conclusion}

In this work, we proposed a transfer-based black-box Energy Attack, and demonstrated its query efficiency. Interestingly, we find that the captured energy distribution of small patches share plenty of similarities among different models and datasets under various settings. Despite the effectiveness of our method, there are some limitations: a) the PCA-based extraction process only captures linear correlations, and only the global distribution of energy, ignoring non-linear or data-dependent aspects; b) we still lack a theoretical analysis on why the distribution similarities arise under different settings, and know little about the larger structures beyond the level of small patches. The high performance of Energy Attack shows a strong potential in this direction. We leave these for future work.

\bibliography{aaai22}

\end{document}